\documentclass[10pt,twocolumn]{wlscirep}
\usepackage[utf8]{inputenc}
\usepackage[T1]{fontenc}
\usepackage{tabularray}
\usepackage{booktabs}
\usepackage{multirow}
\usepackage{amsmath}
\usepackage{algorithm}
\usepackage{algorithmic}

\title{HiCMamba: Enhancing Hi-C Resolution and Identifying 3D Genome Structures with State Space Modeling}

\author[1,$\dag$]{Minghao Yang}
\author[2,$\dag$]{Zhi-An Huang}
\author[3]{Zhihang Zheng}
\author[3]{Yuqiao Liu}
\author[3]{Shichen Zhang}
\author[1]{Pengfei Zhang}
\author[1,*]{Hui Xiong}
\author[3,*]{Shaojun Tang}
\affil[1]{Artificial Intelligence Thrust, Information Hub, Hong Kong University of Science and Technology (Guangzhou), Guangzhou, 511466, China}
\affil[2]{Research Office, City University of Hong Kong (Dongguan), Dongguan 523000, China}
\affil[3]{Bioscience and Biomedical Engineering Thrust, System Hub, Hong Kong University of Science and Technology (Guangzhou), Guangzhou, 511466, China}
\affil[*]{Address correspondence to: xionghui@ust.hk; shaojuntang@ust.hk}
\affil[$\dag$]{These authors contributed equally to this work.}

\begin{abstract}
Hi-C technology measures genome-wide interaction frequencies, providing a powerful tool for studying the 3D genomic structure within the nucleus. However, high sequencing costs and technical challenges often result in Hi-C data with limited coverage, leading to imprecise estimates of chromatin interaction frequencies. To address this issue, we present a novel deep learning-based method HiCMamba to enhance the resolution of Hi-C contact maps using a state space model. 
We adopt the UNet-based auto-encoder architecture to stack the proposed holistic scan block, enabling the perception of both global and local receptive fields at multiple scales.
Experimental results demonstrate that HiCMamba outperforms state-of-the-art methods while significantly reducing computational resources. Furthermore, the 3D genome structures, including topologically associating domains (TADs) and loops, identified in the contact maps recovered by HiCMamba are validated through associated epigenomic features. Our work demonstrates the potential of a state space model as foundational frameworks in the field of Hi-C resolution enhancement.
\end{abstract}
\begin{document}

\flushbottom
\maketitle

\thispagestyle{empty}

\section*{Introduction}
Nuclear genomes house the majority of genetic information essential for determining the phenotype of cells, tissues, and organisms \cite{misteli2020self}. Within the nucleus, chromosomes are intricately folded and organized in three-dimensional space, allowing various chromosomal loci to interact with one another \cite{bonev2016organization, rowley2018organizational}. This 3D genome architecture plays a critical role in regulating gene expression and maintaining cellular homeostasis \cite{zhang2024computational, monteagudo2024impact, cuartero2023three}. High-throughput chromosome conformation capture (Hi-C) \cite{lieberman2009comprehensive} has become a powerful method for measuring the 3D genome structure, enabling the discovery of inherent hierarchical topological features such as A/B compartments \cite{lieberman2009comprehensive}, topologically associating domains (TADs) \cite{dixon2012topological}, and loops \cite{rao20143d}. Low-resolution Hi-C data result in blurred TADs and loops, highlighting the necessity of using high-resolution Hi-C sequencing data to accurately identify these spatial patterns. In recent years, high-resolution Hi-C data (e.g., 10kb and 40kb) \cite{rao20143d, dixon2012topological} have become available, enabling more efficient and accurate identification of TADs and loops. High-resolution Hi-C data are increasingly in demand among researchers investigating the intricate 3D structures of chromosomes. Nevertheless, owing to technical constraints and high sequencing costs, most publicly available high-resolution Hi-C data are derived from labor-intensive and time-consuming experiments. Consequently, there is an urgent need to develop computational methods to improve the resolution of Hi-C data. Advancement of deep-learning methods for in-silico image refinement made improvement of low-resolution HiC data possible. Recently, computational tools have been proposed to expedite the enhancement of HiC data resolution. 

The Hi-C data are generally displayed as an $n \times n$ contact matrix, with the chromosome segmented into $n$ equally sized bins. Specifically, the value of each cell in the matrix reflects the frequency of interaction between two genomic loci. The high resolution of Hi-C contact matrices, often exceeding 10,000 bins for a single human chromosome (e.g., a 100 Mb chromosome with 10kb bin width), poses a significant challenge for current deep learning methods. These methods typically partition the matrices into non-overlapping blocks and downsample the original high-coverage maps, resulting in lower-resolution. Low-coverage Hi-C contact maps, generated through downsampling, are then fed into the deep learning models to reconstruct the original high-coverage maps. Existing computational methods for enhancing low-coverage Hi-C contact matrices can be generally categorized into two groups: the traditional convolutional neural network (CNN)-based and the generative adversarial network (GAN)-based methods. 
Traditional CNN-based methods, such as HiCPlus \cite{zhang2018enhancing} and HiCNN \cite{tong2019hicnn}, employ multiple CNN blocks to predict high-coverage contact maps from low-coverage inputs. However, their reliance on mean square error (MSE) loss often leads to over-smoothed and blurry predictions \cite{li2023ienhance}.
Moreover, several GAN-based models, e.g., HiCSR \cite{dimmick2020hicsr} and HiCARN \cite{hicks2022hicarn}, have been proposed to generate high-coverage contact maps. These methods include a generator that transforms low-coverage contact maps into high-coverage contact maps, and a discriminator that takes both the generated and real contact maps as input, classifying them as either real or fake.

While previous methods for enhancing Hi-C contact maps have shown promise, they suffer from several limitations for their heavy reliance on CNNs. CNNs, with their inherent local receptive fields, struggle to capture the crucial long-range dependencies present in Hi-C data. In addition, these methods, coupled with the need for deep architectures to compensate, often results in high computational complexity and cost. Besides, GAN-based methods often face convergence issues for the adversarial training between the generator and discriminator.

To tackle these challenges, we propose a Mamba-based \cite{gu2023mamba} auto-encoder framework HiCMamba, leveraging state space model \cite{gu2023mamba} to infer the high-coverage Hi-C contact maps based on UNet architecture \cite{ronneberger2015u}. HiCMamba incorporates a novel holistic scan block within each layer to effectively capture multi-scale features. This block consists of a two-dimensional selective scan (SS2D) module and a locally-enhanced feedforward neural network (LEFN). The SS2D module achieves a global receptive field with linear complexity by using a four-way sequential scanning strategy based on SSM.
The LEFN consisting of multi-layer CNN captures local information of neighborhood pixels. 

Extensive experiments on Hi-C datasets demonstrate that HiCMamba outperforms state-of-the-art methods in both effectiveness and generalization while significantly reducing computational cost. Notably, HiCMamba achieves these high-quality recovery results with only 25\% of the computational cost compared to the runner-up method. HiCMamba exhibits global receptive fields, in contrast to other methods that are restricted to local receptive fields. 
Furthermore, HiCMamba-recovered contact maps showcase accurate 3D genome structures identification, such as TADs and loops. Finally, we find that the 3D genome structure, along with epigenomic features such as ChIP-Seq, methylation, and super-enhancer (SE) regions, plays an integral role in regulating gene expression.

\begin{figure*}[t]
\begin{center}
\includegraphics[width=0.91\linewidth]{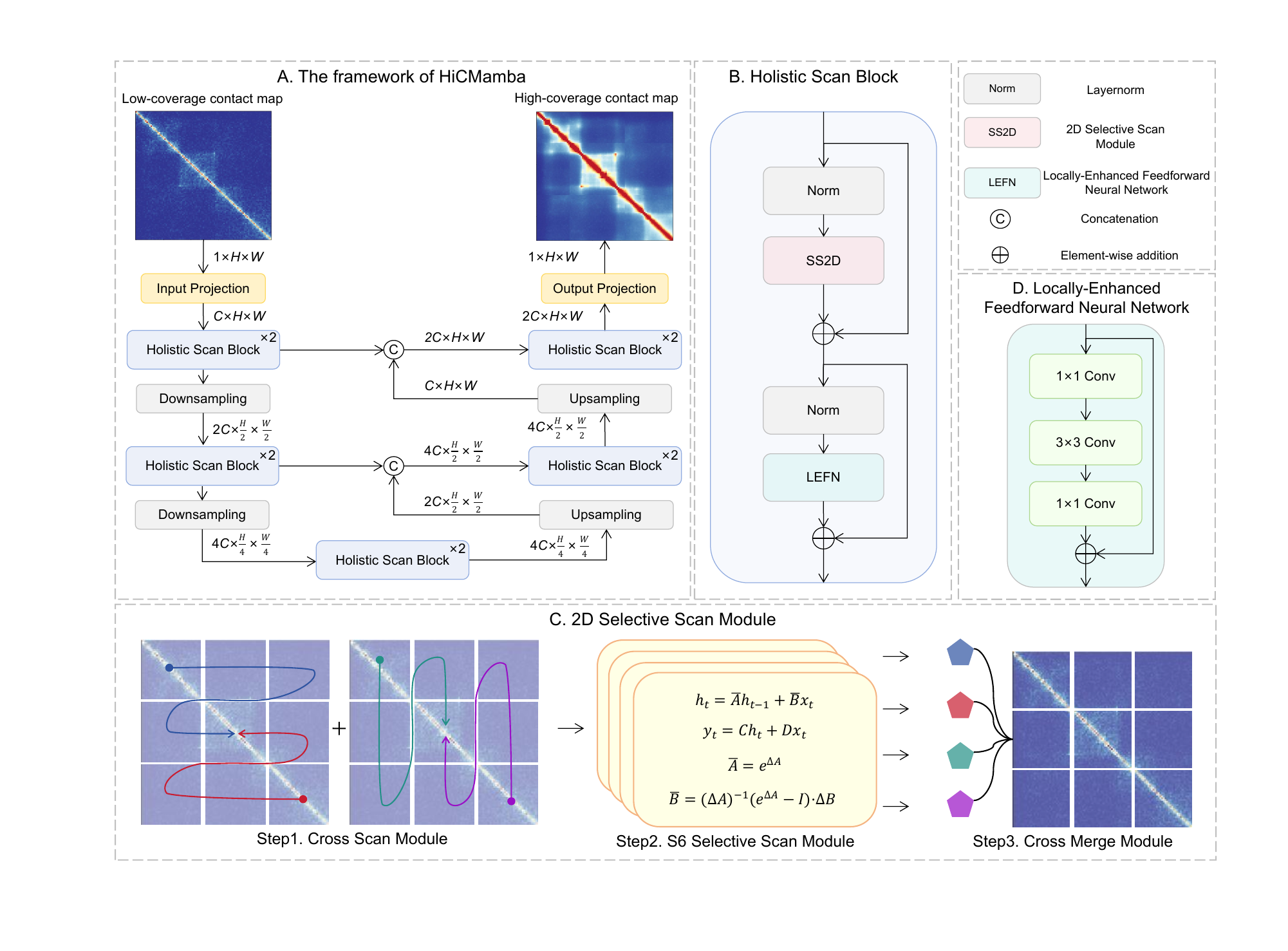}
\end{center}
\caption{
\textbf{Overview of HiCMamba algorithm.} (\textbf{A}) The framework of HiCMamba. The workflow begins with an input projection layer that extracts shallow features from the low-coverage Hi-C contact map (input). These extracted features are then input into a UNet-based auto-encoder architecture, which utilizes our proposed holistic scan block. This block facilitates the feature extraction and reconstruction at multiple scales. In the final stage, these refined features pass through an output projection layer to reconstruct the final high-coverage Hi-C contact map. (\textbf{B}) The architecture of the proposed holistic scan block, structured as Norm $\rightarrow$ SS2D $\rightarrow$ Norm $\rightarrow$ LEFN, following the design of the Transformer block. (\textbf{C}) Diagram of the SS2D module. First, the input features are flattened along four distinct scanning paths. Each path is processed independently by an individual S6 selective scan module. Finally, the outputs from each path are combined to reconstruct the 2D feature map. (\textbf{D}) Structure illustration of LEFN.
}
\label{independent}
\end{figure*}

\section*{Materials and Methods}
\subsection*{Hi-C Datasets and data preprocessing}
The raw Hi-C data consists of an $n \times n$ interaction frequency contact matrix, representing the all-versus-all interaction mapping of fragments within a chromosome \cite{lieberman2009comprehensive}. Each matrix entry indicates the interaction frequency between a pair of fragments, where $n$ represents the number of fragments in a chromosome at a given Hi-C data resolution.

In this study, the high-coverage Hi-C data of 10kb resolution are downloaded from GEO GSE63525. (\url{https://www.ncbi.nlm.nih.gov/geo/query/acc.cgi?acc=GSE63525}). Two widely investigated cell lines (i.e., GM12878 and K562) are employed to assess the effectiveness of HiCMamba. Following previous works \cite{hicks2022hicarn, hong2020deephic}, we preprocess the 10-kb resolution Hi-C data through normalization, down-sampling, and data division. Specifically, KR normalization \cite{knight2013fast} is applied to the high-resolution Hi-C contact maps, derived from paired-end sequencing reads with a mapping quality greater than 30. Low-coverage contact maps are then generated by downsampling the normalized data at a ratio of $1/16$, simulating the lower resolutions achieved at reduced sequencing depths in practice \cite{zhang2018enhancing}. The contact maps are subsequently split into non-overlapping $40 \times 40$ sub-matrices. 
Low-coverage and high-coverage (target) contact maps are created by concatenating these sub-matrices, with the high-coverage maps skipping the downsampling step.
For data partitioning, we follow the approach of previous studies \cite{hicks2022hicarn}: chromosomes 2, 6, 10, and 12 are used for validation; chromosomes 4, 14, 16, and 20 for testing, and the remaining chromosomes for training.

\subsection*{The HiCMamba algorithm}
In this work, our aim is to enhance the resolution of Hi-C data given a low-resolution contact map.
Here, we develop a multi-scale framework HiCMamba, leveraging the state space model \cite{gu2023mamba} within a hierarchical UNet framework to capture long-range dependencies while reducing computational costs. 
As shown in Figure~1A, HiCMamba utilizes a UNet-based auto-encoder architecture that incorporates holistic scan blocks to extract features across multiple scales. 
Specifically, the proposed HiCMamba comprises an input projection layer, an encoder, a bottleneck layer, a decoder, and an output projection layer. Convolutional layers in the input and output projection layers extract the low-level features and reconstruct high-coverage contact maps, respectively. Two holistic scan blocks, for the extraction of global and local representations using SS2D and LEFN, are used in the bottleneck layer and in each layer of the encoder and decoder.
Given a single channel low-coverage Hi-C contact map $X_{\rm in} \in \mathbb{R}^{1 \times H \times W}$, a convolutional layer is first applied in the input projection layer to extract the shallow representations $E_1 \in \mathbb{R}^{C \times H \times W}$. $C$ represents the predefined feature dimension, while $H$ and $W$ denote the height and width of the input contact map, respectively. Next, $E_1$ is iteratively passed through two layers of holistic scan blocks along with downsampling, resulting in $E_2 \in \mathbb{R}^{2C \times \frac{H}{2} \times \frac{W}{2}}$ and $E_3 \in \mathbb{R}^{4C \times \frac{H}{4} \times \frac{W}{4}}$. A bottleneck stage composed of holistic scan blocks is incorporated at the end of the encoder. For feature reconstruction, the proposed decoder also comprises two stages, with upsampling following holistic scan blocks. The features $E_i$ from the $i$-th encoder stage are concatenated with $D_i$ from the previous decoder stage using skip connections. Where $D_i$ is the output of the $i$-th decoder stage with the same shape as $E_i$. Finally, the output of the decoder is fed into the output projection layer to obtain the high-cover contact map. 

Figure~1B showcases the backbone of the holistic scan block, structured as Norm $\rightarrow$ SS2D $\rightarrow$ Norm $\rightarrow$ LEFN, similar to the design of the Transformer block \cite{vaswani2017attention}. Here, $\text{Norm}$ denotes layer normalization \cite{ba2016layer}. The SS2D module (Figure~1C) enables a comprehensive scan of information from different directions with low time complexity, and the LEFN module (Figure~1D) leverages multiple CNNs to facilitate the inception of a local receptive field. The combination of SS2D and LEFN effectively captures both global and local receptive fields simultaneously. The details of the holistic scan block, the SS2D module, and the LEFN module are described in the following section.

HiCMamba is developed using Python and PyTorch, and executed on the Ubuntu platform with a Tesla A100 GPU. HiCMamba handles the low-resolution input at a $40\times40$ resolution. The number of holistic scan blocks in the bottleneck layer, as well as in each layer of the encoder and decoder, is set to two. The predefined feature dimension $C$ is set to 32. The number of neurons of input features of bottleneck layer, the $i$-th layer of encoder, and the $i$-th layer of decoder are $4C$, $i \times C$, and $i \times 2C$, respectively. The LEFN is composed of three layers of CNNs with kernel sizes of $1 \times 1$, $3 \times 3$, and $1 \times 1$. The filters for the bottleneck layer, the $i$-th layer of encoder, and $i$-th layer of decoder are $\in \{4C, 4C, C\}$, $\in \{i \times 4C, i \times  4C, i \times  C\}$, and $\in \{i \times 8C, i \times  8C, i \times  2C\}$, respectively. We set the batch size to 64 and train HiCMamba using the Adam optimizer \cite{kingma2014adam}, with a learning rate of 1e-4 and momentum parameters $\beta_1$ and $\beta_2$ set to 0.9 and 0.999, respectively.
The L1 loss is adopted for training, which has been demonstrated to be less susceptible to over-smoothing \cite{zhang2023reference}. The definition of L1 loss is provided as follows:
\begin{equation}
\text{L1}(\hat{y}, y) = \frac{1}{mn} \sum_{i=0}^{m-1} \sum_{j=0}^{n-1} |\hat{y}(i,j) - y(i,j)|
\end{equation}

\paragraph*{Mamba-based Holistic Scan Block}
We propose a novel holistic scan block (Figure 1B) to address the limitations of traditional CNN and GAN-based methods for Hi-C resolution enhancement. 

First, the SS2D module within the block effectively captures long-range dependencies in Hi-C data. This is achieved by a multi-path scanning strategy coupled with robust sequential modeling capabilities of the S6 block. By gathering information from multiple directions and leveraging the S6 block’s advanced sequence modeling, the holistic scan block accurately represents distant genomic loci interactions, ensuring comprehensive global information extraction.

Second, unlike computationally expensive deep CNN structures, our approach leverages the linear time complexity of the state space model for efficiency. The Mamba-based holistic scan block performs efficient state transitions, requiring significantly less computational power compared to deep convolutional layers. This inherent efficiency eliminates the need for complex, computationally demanding deep architectures.

Third, by avoiding adversarial training, HiCMamba overcomes the convergence challenges often encountered in GAN-based methods. The robust modeling capabilities of SS2D and LEFN within the holistic scan block ensure stable and consistent performance throughout the training process.

Moreover, state space model does not explicitly compute dependencies among pixels and is thus not proficient in capturing local information \cite{deng2024facing}. To recognize the importance of local context for accurate Hi-C resolution enhancement, our holistic scan block incorporates the LEFN module. This module complements the global dependency capture of the SS2D module, enabling detailed reconstruction of high-resolution contact maps by leveraging information from neighboring pixels. 

We structure the holistic scan block following the design of the Transformer \cite{vaswani2017attention} block, improving gradient flow and training stability by applying layer normalization before the SS2D and LEFN. Overall, the proposed holistic scan block effectively captures both long-range dependencies and valuable local context by combining the strengths of state space modeling and convolutional operations. The SS2D and LEFN components are detailed in the following subsections.

\begin{figure*}[t]
\begin{center}
\includegraphics[width=0.85\linewidth]{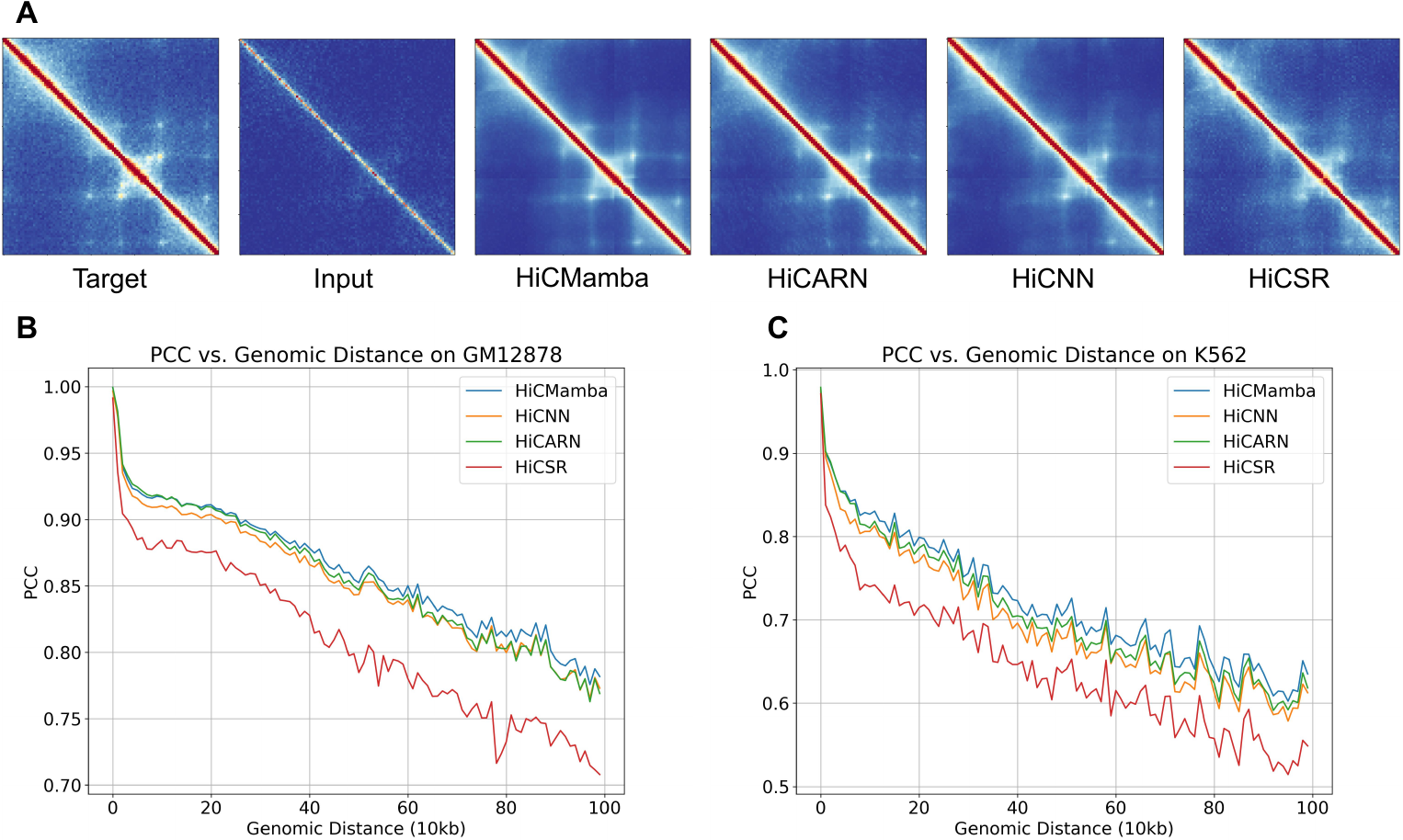}
\end{center}
\caption{
\textbf{HiCMamba enhances the contact matrix.} (\textbf{A}) 
Contact maps for a 1-Mb genomic region of Chromosome 14 (32Mb-33Mb) from the GM12878 dataset. The first column represents the full-coverage map, the second column represents the low-coverage input, the third column shows the enhanced map generated by HiCMamba, and the remaining columns display maps from compared methods.
(\textbf{B}) Performance evaluation of HiCMamba and other models across various genomic distances. The Pearson Correlation Coefficient (PCC) between enhanced contact maps and the actual high-resolution contact maps is calculated for both the GM12878 (left panel) and K562 (right panel), respectively.
}
\end{figure*}

\begin{table*}[ht]
\caption{Performance evaluation of HiCMamba compared to state-of-the-art methods using the GM12878 and K562 datasets. Results highlighted in \textbf{bold} and \underline{underlined} represent the best and second-best, respectively. Subscript indicates the percentage improvement of our method compared to the second-best method. The formula is given by: 
$\text{Improvement (\%)} = \frac{\text{A} - \text{B}}{\text{B}} \times 100\%$, where A and B are the best and the second-best method, respectively.
}
\setlength{\tabcolsep}{5mm}{
\begin{tabular}{@{}clllll@{}}
\toprule
Datasets                                     & Methods         & SSIM $\uparrow$   & PSNR $\uparrow$    & PCC $\uparrow$    & SRCC $\uparrow$   \\ \midrule
\multicolumn{1}{c}{\multirow{4}{*}{GM12878}} & HiCSR           & 0.8969 & 33.2447 & 0.5573 & 0.5050 \\
\multicolumn{1}{c}{}                         & HiCNN           & 0.9104 & 34.7939 & 0.5952 & 0.5322 \\
\multicolumn{1}{c}{}                         & HiCARN          & \underline{0.9111} & \textbf{35.1002} & \underline{0.5867} & \underline{0.5242} \\
\multicolumn{1}{c}{}                         & HiCMamba (Ours) & $\textbf{0.9141}_{\textcolor{green}{+0.33\%}}$ & $\underline{35.0686}_{\textcolor{red}{-0.09\%}}$ & $\textbf{0.6113}_{\textcolor{green}{+4.19\%}}$ & $\textbf{0.5497}_{\textcolor{green}{+4.86\%}}$ \\ \midrule
\multirow{4}{*}{K562}                        & HiCSR           & 0.9356 & 36.9395 & 0.4288 & 0.3416 \\
                                             & HiCNN           & 0.9455 & 38.5005 & 0.4963 & 0.3918 \\
                                             & HiCARN          & \underline{0.9478} & \underline{38.7629} & \underline{0.5042} & \underline{0.3955} \\
                                             & HiCMamba (Ours) & $\textbf{0.9487}_{\textcolor{green}{+0.09\%}}$ & $\textbf{38.8100}_{\textcolor{green}{+0.12\%}}$ & $\textbf{0.5219}_{\textcolor{green}{+3.51\%}}$ & $\textbf{0.4136}_{\textcolor{green}{+4.58\%}}$ \\ \bottomrule
\end{tabular}}
\end{table*}

\paragraph*{Global Receptive Fields via 2D Selective Scan}
Inspired by state space model \cite{liu2024vmamba}, we introduce the 2D Selective Scan (SS2D) module to effectively capture global receptive fields, ensuring comprehensive and efficient feature extraction from the genomic interaction data.
As illustrated in Figure~1C, SS2D comprises three steps: cross scan module, S6 selective scan module \cite{gu2023mamba}, and cross merge module. 

First, the cross scan module initially transforms the input contact map into sequences along four unique traversal paths (i.e., from top-left to bottom-right, bottom-right to top-left, top-right to bottom-left, and bottom-left to top-right). 
This method is particularly suited to the complex nature of Hi-C data, ensuring a thorough scan of interaction frequencies. The generated sequences are then fed into the S6 selective scan module for a detailed representation of the contact map.

Second, the S6 selective scan module, a variant of the state space model with a selective scan mechanism, functions as a linear time-invariant system.
Mathematically, it maps the input state $x(t)$ to the output state $y(t)$ via the hidden state $h(t)$, which is typically represented by linear ordinary differential equations as follows:
\begin{align}
h'(t) &= Ah(t) + Bx(t) \\
y(t) &= Ch(t)
\end{align}
where $A$, $B$, and $C$ are learnable parameter matrices. $h'(t)$ represents the derivative of the hidden state $h(t)$ at time step $t$. $A$ retains historical information, shaping the influence of the prior hidden state on the current hidden state, while $B$ quantifies the impact of the input $x(t)$ on the hidden state. $C$ delineates the transformation of the hidden state into the output. 

To be incorporated into deep learning models, continuous-time state space models must be discretized beforehand, which can be obtained using the zeroth-order hold method as follows:
\begin{align}
h_t &= \overline{A} h_{t-1} + \overline{B} x_t \\
y_t &= C h_t\\
\overline{A} &= e^{\Delta A} \\
\overline{B} &= (\Delta A)^{-1} (e^{\Delta A} - I) \cdot \Delta B
\end{align}
where $h_t$ is the discrete hidden state at time step $t$, while $h_{t-1}$ represents the hidden state of the previous step. The discrete input and output representation at time step $t$ are denoted as $x_t$ and $y_t$, respectively. The continuous parameters $A$ and $B$ are converted to discrete parameters $\overline{A}$ and $\overline{B}$ using the zeroth-order hold method with a timescale parameter $\Delta$. 

The iterative calculation of $y$ can be accelerated using parallel global convolutional operation as follows:
\begin{align}
y &= x \odot \overline{K} \\
\overline{K} &= (C\overline{B}, C\overline{A}\overline{B}, \ldots, C\overline{A}^{T-1}\overline{B})
\end{align}
where $\odot$, $\overline{K}$, and $T$ represent the global convolutional operation, the convolutional kernel, and the total number of pixels, respectively. The pseudo code for S6 selective scan module is provided in Algorithm~1.

Finally, the cross merge module combines and merges the sequence representations extracted by the four S6 modules to reconstruct the high-coverage contact map, maintaining the same size as the input low-coverage contact map. 

The adaptation of SS2D to Hi-C data is crucial for addressing the challenges posed by genomic interactions, which enables HiCMamba to accurately capture long-range dependencies between pixels and complex chromatin interactions with high efficiency. This advancement improves the resolution of Hi-C contact maps with reduced computational costs, providing a powerful tool for 3D genomic data analysis.

\begin{algorithm}
\caption{Pseudo code for S6 selective scan module}
\begin{algorithmic}[1]

\renewcommand{\algorithmicrequire}{\textbf{Input:}}
\renewcommand{\algorithmicensure}{\textbf{Output:}}
\REQUIRE $x$, the raw feature maps
\ENSURE $y$, the refined feature maps
\STATE \textbf{Params:} $W_{A}$, $W_{B}$, $W_C$, the learnable parameters for linear transformations
\STATE \textbf{Params:} $\Delta$, the timescale parameter for discretization

\STATE \textbf{Step 1: Linear Transformations}
\STATE $A, B, C \gets \text{Linear}(x, W_{A}), \text{Linear}(x, W_{B}), \text{Linear}(x, W_C)$

\STATE \textbf{Step 2: Exponential Calculations and Discretization}
\STATE $\overline{A} \gets \exp(\Delta \cdot A)$
\STATE $\overline{B} \gets (\Delta \cdot A)^{-1} \cdot (\exp(\Delta \cdot A) - I) \cdot B$

\STATE \textbf{Step 3: Global Convolutional Operation for Acceleration}
\STATE $\overline{K} \gets (C\overline{B}, C\overline{A}\overline{B}, \ldots, C\overline{A}^{T-1}\overline{B})$
\STATE $y \gets x \odot \overline{K}$

\RETURN $y$
\end{algorithmic}
\end{algorithm}

\paragraph*{Local Receptive Fields via Locally-Enhanced Feedforward Neural Network}
Previous research \cite{wu2021cvt, wang2022uformer} has highlighted the limited ability of feedforward neural networks to utilize local context effectively. Given the importance of neighboring pixels in Hi-C contact map recovery, LEFN is designed to enable a local receptive field. As shown in Figure~1D, we first apply a $1 \times 1$ convolution layer to each token to enhance the feature dimension. A $3 \times 3$ convolution layer is then used to capture local information. Finally, the features are processed through another $1 \times 1$ convolution layer to reduce the channels for matching match the input dimension. Each CNN layer is followed by a GELU activation layer \cite{hendrycks2016gaussian}.

\subsection*{Loop weighted scoring}
We propose a loop weighted score to measure the correlation between cell type-specific loops and cell type-specific SEs. First, the proportion of SE-related loops for each cell line is calculated as follows:
\begin{equation}
P_l^s = \frac{A_l^s}{N_l}
\end{equation}
where loop-related and SE-related cell lines are denoted as $l \in \{ {\rm GM12878}, {\rm K562} \}$ and $s \in \{ {\rm GM12878}, {\rm K562} \}$, respectively. $A_l^s$ is the number of $l$-specific loops associated with the $s$-specific SEs, and $N_l$ represents the total number of loops identified in cell line $l$. Then, we calculate the loop weighted score $W_l^s$ is as follows:
\begin{equation}
W_l^s = \frac{P_l^s}{P_{\rm GM12878}^s + P_{\rm K562}^s}
\end{equation}
This score represents the relative contribution of each cell line's SE-related loops to the total SE-related loops across both cell lines.

\subsection*{Evaluation metrics}
To comprehensively compare HiCMamba with other state-of-the-art methods, we employ four key metrics: structural similarity index measure (SSIM), peak signal-to-noise ratio (PSNR), Pearson correlation coefficient (PCC), and Spearman rank correlation coefficient (SRCC).
SSIM assesses the similarity between two images from multiple perspectives, including luminance, structure, and contrast. The calculation of SSIM can be reached as follows:
\begin{equation}
\text{SSIM}(\hat{y}, y) = \frac{(2\mu_{\hat{y}}\mu_{y} + C_1)(2\sigma_{\hat{y}y} + C_2)}{(\mu_{\hat{y}}^2 + \mu_{y}^2 + C_1)(\sigma_{\hat{y}}^2 + \sigma_{y}^2 + C_2)}
\end{equation}
where $\mu_{\hat{y}}$ and $\mu_{y}$ represent the mean values of the predicted contact map $\hat{y}$ and the ground-truth high-coverage contact map $y$, respectively. $\sigma_{\hat{y}}^2$, $\sigma_{y}^2$, and $\sigma_{\hat{y}y}$ are their respective variances. The constants $C_1$ and $C_2$ are typically set to $0.01^2$ and $0.03^2$, respectively. 

PSNR quantifies the ratio of the maximum possible power of a signal to the power of noise that affects the signal quality. PSNR is formulated as follows:

\begin{align}
&\text{PSNR}(\hat{y}, y) = 10 * \log_{10} \left( \frac{\text{MAX}^2}{\text{MSE}{(\hat{y}, y)}} \right) \\
&\text{MSE}(\hat{y}, y) = \frac{1}{mn} \sum_{i=0}^{m-1} \sum_{j=0}^{n-1} [\hat{y}(i,j) - y(i,j)]^2
\end{align}
where \text{MAX} is the maximum pixel value of the contact map, set to 1 due to normalization. $\hat{y}(i,j)$ and $y(i,j)$ indicate the pixel values at position $(i,j)$ in predicted and ground-truth contact map, respectively. The variables $m$ and $n$ denote the number of rows and columns in the contact map, respectively. 
To measure the linear correlation, PCC can be obtained as follows:
\begin{equation}
\text{PCC}(\hat{y}, y) = \frac{\sum_{i=1}^n (\hat{y}_i - \mu_{\hat{y}})(y_i - \mu_{y})}{\sqrt{\sum_{i=1}^n (\hat{y}_i - \mu_{\hat{y}})^2} \sqrt{\sum_{i=1}^n (y_i - \mu_{y})^2}}
\end{equation}
To assesses monotonic relationships, SRCC can be written as:
\begin{equation}
\text{SRCC}(\hat{y}, y) = \frac{\sum_{i=1}^n (R(\hat{y}_i) - \overline{R}(\hat{y}))(R(y_i) - \overline{R}(y))}{\sqrt{\sum_{i=1}^n (R(\hat{y}_i) - \overline{R}(\hat{y}))^2} \sqrt{\sum_{i=1}^n (R(y_i) - \overline{R}(y))^2}}
\end{equation}
where $R(\hat{y}_i)$ and $R(y_i)$ denote the ranks of the $i$-th pixel values in the predicted contact map $\hat{y}$ and the ground-truth contact map $y$, respectively. $\overline{R}(\hat{y})$ and $\overline{R}(y)$ are their respective mean ranks.

\begin{figure*}[t]
\begin{center}
\includegraphics[width=0.9\linewidth]{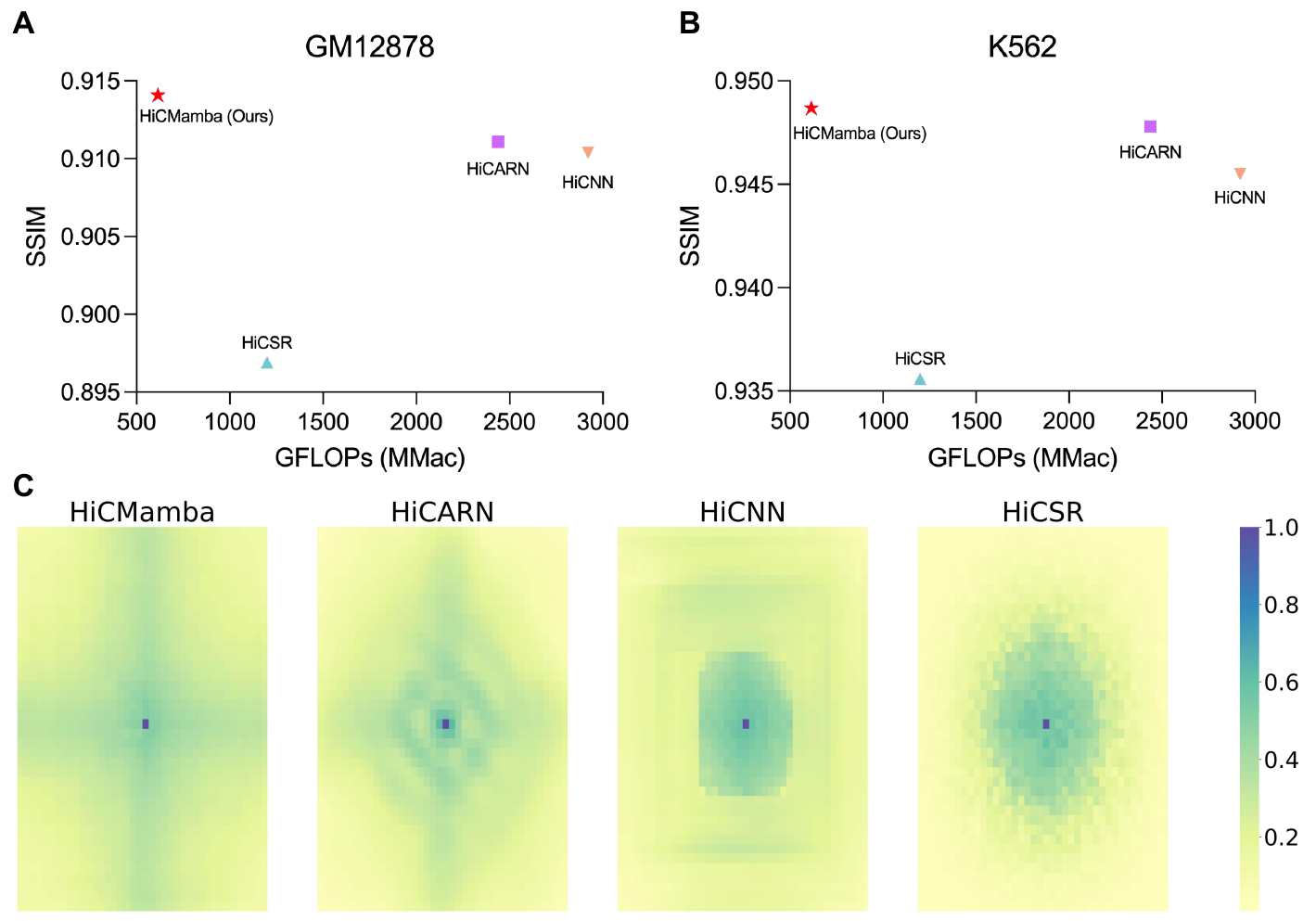}
\end{center}
\caption{
\textbf{Performance, computational cost, and receptive field comparison of HiCMamba with state-of-the-art methods.} (\textbf{A-B}) SSIM versus the computational cost of HiCMamba and alternative tools on the GM12878 (A) and K562 (B) datasets, respectively. (\textbf{C}) Visualization of the effective receptive field (ERF) for HiCMamba compared to other state-of-the-art methods. 
}
\end{figure*}

\section*{Results}

\subsection*{HiCMamba demonstrates efficacy in recovering high-coverage Hi-C contact maps.}
HiCMamba is benchmarked with three state-of-the-art methods, involving HiCNN \cite{tong2019hicnn}, HiCSR \cite{dimmick2020hicsr}, and HiCARN \cite{hicks2022hicarn}. Evaluations are conducted on the preprocessed datasets from GM12878 and K562 cell lines. To ensure a fair comparison, all methods are implemented using their default parameters as reported in their respective publications. 

Table~1 illustrates the test performance of the compared methods on GM12878 and K562 datasets, respectively. HiCMamba achieves the highest performance across all four metrics on the K562 and three out of four metrics on the GM12878. 
For GM12878, HiCMamba outperforms the runner-ups by significant margins: 0.19\% in SSIM, -0.04\% in PSNR, 3.32\% in PCC, and 4.14\% in SRCC. For K562, HiCMamba surpasses the second-best method by 0.27\% in SSIM, 2.71\% in PSNR, 4.62\% in PCC, and 4.94\% in SRCC.

Figure~2A visually compares the full-coverage (target), low-coverage (input), and enhanced contact maps predicted by each compared method for a 1-Mb genomic region (chr14:32Mb–33Mb) on the GM12878 dataset. All enhanced contact maps show improvement over the low-coverage input, with HiCMamba and HiCARN exhibiting a greater ability to capture fine-scale structures such as loops. 
Figure~2B and Figure~2C further highlight HiCMamba's superior performance by illustrating the PCCs between predicted and ground-truth high-coverage contact maps for both datasets across various distance ranges. HiCMamba consistently outperforms existing methods, particularly in sparse regions of the contact map.
Overall, the experimental results demonstrate the effectiveness of HiCMamba.
The combination of the UNet architecture, state space models, and locally enhanced feedforward networks allows for efficient capture of both global and local features at multiple scales, leading to superior performance in Hi-C contact map enhancement.

\subsection*{HiCMamba offers a global receptive field with lower time cost.}
In this section, we evaluate the computational efficiency of HiCMamba and other state-of-the-art methods using giga floating point operations per second (GFlops) as a measure of resource utilization.
As depicted in Figure~3A and Figure~3B, HiCMamba achieves the highest SSIM while using only 25\%, 21\%, and 61\% of the GFlops compared to HiCARN, HiCNN, and HiCSR, respectively. 
This highlights HiCMamba's ability to surpass the accuracy of existing methods with greater efficiency.
Beyond computational efficiency, we analyze the effective receptive field \cite{ding2022scaling} of each method, which represents the region within the input space that influences the activation of a specific output unit. 
Focusing on the central pixel, Figure~3C illustrates a key distinction: HiCMamba exhibits global receptive fields, while the other methods display only local receptive fields. 
Although HiCARN theoretically allow for global coverage with its deep convolutional layers, this comes at the expense of a quadratic increase in computational cost.
In contrast, all pixels are engaged in HiCMamba to highlight cross-pixel activation. Furthermore, the integration of the 2D selective scan mechanism and locally-enhanced feedforward neural network together ensure that the central pixel is primarily influenced by pixels along the cross, thereby facilitating both global and local dependency contexts.

\begin{table*}[ht]
\caption{Cross-cell line performance evaluation of Hi-C resolution enhancement methods using GM12878 and K562 as independent test sets alternatively. Results highlighted in \textbf{bold} and \underline{underlined} indicate the best and runner-up, respectively. The subscript denotes the percentage improvement of our method compared to the second-best method, as described in the caption of Table 1.}
\centering
\setlength{\tabcolsep}{1.7mm}{
\begin{tabular}{@{}p{5cm}lllll@{}}
\toprule
Datasets                                     & Methods         & SSIM $\uparrow$   & PSNR $\uparrow$    & PCC $\uparrow$    & SRCC $\uparrow$   \\ \midrule
\multirow{4}{*}{\shortstack[l]{Trained on K562 and \\ tested on GM12878}} & HiCSR           & 0.8773 & 29.9254 & 0.5273 & 0.4644 \\
                                             & HiCNN           & 0.9037 & 32.2952 & 0.5832 & 0.5229 \\
                                             & HiCARN          & \underline{0.9089} & \textbf{33.6074} & \underline{0.5847} & \underline{0.5232} \\
                                             & HiCMamba (Ours) & $\textbf{0.9106}_{\textcolor{green}{+0.19\%}}$ & $\underline{33.5943}_{\textcolor{red}{-0.04\%}}$ & $\textbf{0.6041}_{\textcolor{green}{+3.32\%}}$ & $\textbf{0.5449}_{\textcolor{green}{+4.14\%}}$ \\ \midrule
\multirow{4}{*}{\shortstack[l]{Trained on GM12878 and \\ tested on K562}} & HiCSR           & 0.9343 & 32.8220 & 0.4349 & 0.3693 \\
                                             & HiCNN           & 0.9384 & 32.0382 & 0.4883 & 0.3958 \\
                                             & HiCARN          & \underline{0.9424} & \underline{33.2697} & \underline{0.4911} & \underline{0.3929} \\
                                             & HiCMamba (Ours) & $\textbf{\textbf{0.9449}}_{\textcolor{green}{+0.27\%}}$ & $\textbf{34.1701}_{\textcolor{green}{+2.71\%}}$ & $\textbf{0.5138}_{\textcolor{green}{+4.62\%}}$ & $\textbf{0.4123}_{\textcolor{green}{+4.94\%}}$ \\ \bottomrule
\end{tabular}}
\end{table*}

\subsection*{HiCMamba shows generalization capability across different cell lines}
To assess the real-world generalization capability of HiCMamba, we conduct cross-dataset validation using GM12878 and K562 cell lines. Specifically, the compared models are trained on one cell line and tested on the other, with results represented in Table~2. As expected, all methods showcase a slight performance decrease compared to the within-dataset evaluation (Table~1). However, HiCMamba consistently outperforms other models on the GM12878 cell line and maintains competitiveness on the K562 cell line, demonstrating its robust generalization capability for real-world applications.

\begin{figure*}[t]
\begin{center}
\includegraphics[width=0.9\linewidth]{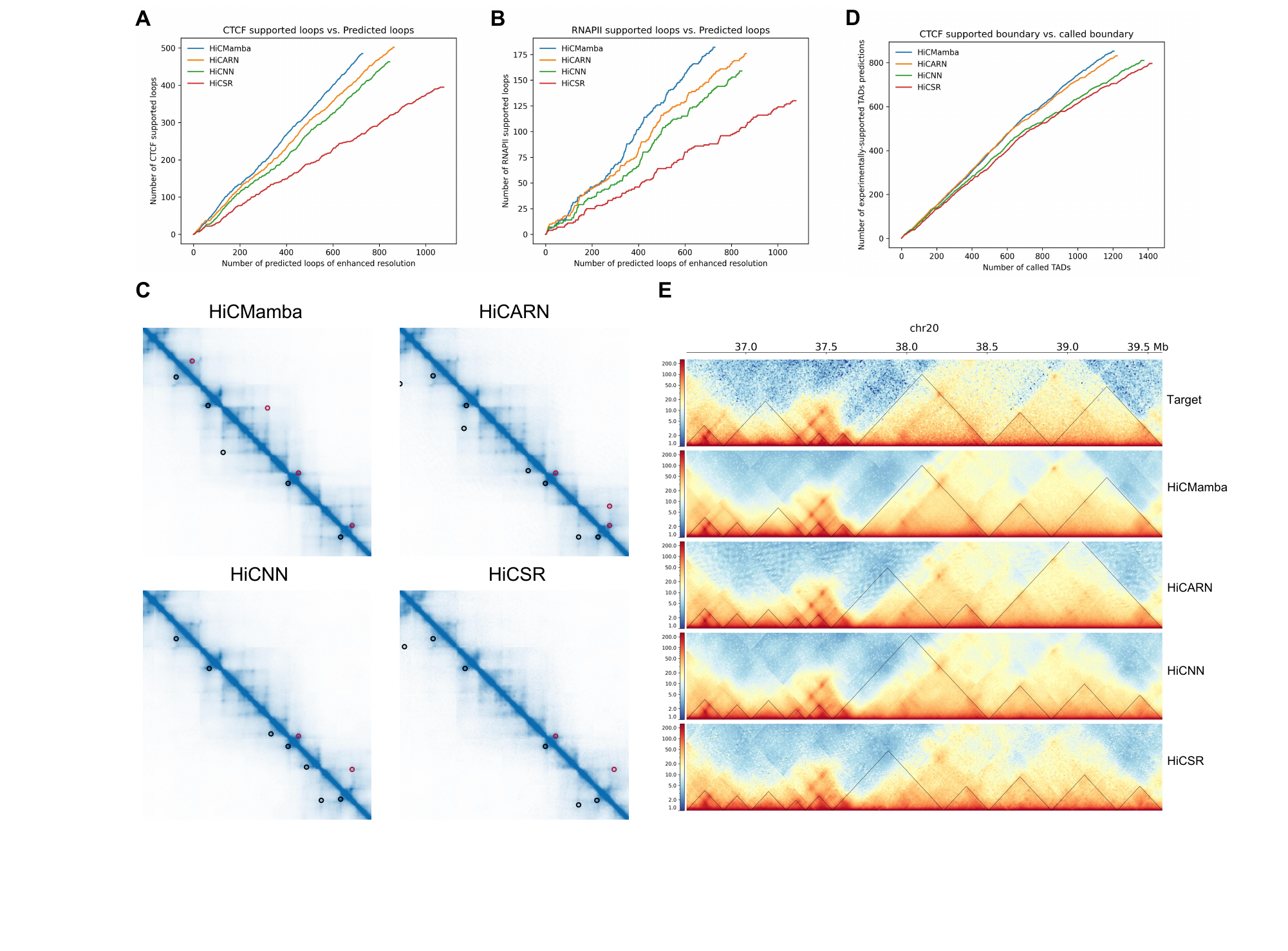}
\end{center}
\caption{
\textbf{Comparison of chromatin loop and TAD annotations derived from enhanced contact maps by HiCMamba and compared methods.} (\textbf{A-B}) 
Comparison of the number of predicted loops on enhanced contact maps against the number of CTCF-supported (A) and RNAPII-supported (B) loops identified through ChIA-PET.
(\textbf{C}) Example of predicted loop annotations (black points , lower left) and ChIA-PET-supported loops (red points, upper right) within a region of Chromosome 16: 67Mb-69Mb from the GM12878 dataset. (\textbf{D}) 
Comparison of the number of predicted TAD boundaries on enhanced contact maps against the number of CTCF-supported TAD boundaries.
(\textbf{E}) Example of TAD boundaries identified from the original high-coverage HiC map (top panel), and enhanced contact maps generated by four deep learning methods (botton four panels)  within a region of Chromosome 20: 37.5-39.5Mb from the GM12878 dataset.
}
\end{figure*}

\subsection*{HiCMamba enhances chromatin loop detection}
Accurate chromatin loop identification is crucial for understanding gene regulation and disease mechanisms \cite{oberbeckmann2024vitro, tam2024cell}. This section investigates whether HiCMamba enhances chromatin loop detection. We employ the off-the-shelf loop annotation tool HiCCUPS \cite{rao20143d} to contact maps generated by HiCMamba and other deep learning methods using the GM12878 dataset. We compared the predicted loops against the ground-truth set of CTCF and RNAPII-supported loops derived from the Chromatin Interaction Analysis by Paired-End Tag Sequencing (ChIA-PET) \cite{tang2015ctcf} data (GSE72816). Table~3 and Figure~4A and Figure~4B showcase the number of predicted loops and their overlap with ChIA-PET-validated loops. HiCMamba outperforms the runner-up with a precision improvement of 8.4\% for CTCF-supported loops and 4.6\% for RNAPII-supported loops. Figure~4C provides a visual comparison in the region of chromosome 16: 67Mb-69Mb, highlighting the minimal false positive rate achieved by HiCMamba.
Black points (lower left) represent loops detected on enhanced contact maps, while red points (upper right) denote the transcription factor-supported loops validated by ChIA-PET. These results suggest that HiCMamba generates more fine-grained contact maps, which are crucial for accurate chromatin loop identification.

\begin{table}
\caption{Comparison of CTCF and RNAPII-supported ChIA-PET loops with loops detected in enhanced contact maps generated by various deep learning methods.}
\centering
\begin{tblr}{
colsep=1.5mm,
  cell{2}{1} = {r=4}{},
  cell{6}{1} = {r=4}{},
  hline{1-2,6,10} = {-}{},
}
                 & Methods  & Matched & Predicted & Proportion \\
CTCF   & HiCMamba & 485     & 730       & 66.4\%     \\
                 & HiCARN   & 502     & 865       & 58.0\%     \\
                 & HiCNN    & 463     & 846       & 54.7\%     \\
                 & HiCSR    & 395     & 1079      & 36.6\%     \\
RNAPII & HiCMamba & 182     & 730       & 24.9\%     \\
                 & HiCARN   & 176     & 865       & 20.3\%     \\
                 & HiCNN    & 159     & 846       & 18.8\%     \\
                 & HiCSR    & 130     & 1079      & 12.0\%     
\end{tblr}
\end{table}

\begin{table}
\caption{Comparison of CTCF ChIP-Seq supported boundaries with those detected in enhanced contact maps generated by deep learning methods.}
\centering
\begin{tblr}{
    colsep=3mm,
  hline{1-2,6} = {-}{},
}
Methods  & Matched & Predicted & Proportion \\
HiCMamba & 852     & 1207      & 70.6\%     \\
HiCARN   & 831     & 1226      & 67.7\%     \\
HiCNN    & 810     & 1376      & 58.9\%     \\
HiCSR    & 796     & 1422      & 56.0\%     
\end{tblr}
\end{table}

\subsection*{HiCMamba is more accurate in TAD boundary annotation}
Topologically associating domains (TADs) are chromosomal regions formed by chromatin loop extrusion, with boundaries demarcated by architectural proteins \cite{beagan2020existence}. These structures are essential for pinpointing functionally relevant sub-regions such as subTADs and microTADs. To assess the utility of HiCMamba in TAD annotation, we use the hicFindTADs method from HiCExplorer \cite{wolff2020galaxy} to reconstruct the TADs boundaries on enhanced contact maps generated by HiCMamba and benchmarking tools on GM12878. CTCF ChIP-Seq peaks are utilized to measure the precision of the recovered TADs \cite{zhang2022reference}. Table~4 and Figure~4D compare the number of predicted TADs and the number of CTCF ChIP-Seq supported boundaries across different methods. HiCMamba demonstrates superior performance with the highest precision values among all evaluated methods, and surpasses HiCARN, HiCNN, and HiCSR by 2.9\%, 11.7\%, and 14.6\%, respectively. 

Figure~4E provides a visual comparison of predicted TAD boundaries and CTCF ChIP-seq supported boundaries within the region of chromosome 20: 36.5Mb-39.5Mb.
Although other methods demonstrate partial TAD reconstruction, HiCMamba consistently identifies more accurate TAD boundaries. 
This enhanced accuracy highlights the effectiveness of HiCMamba in recovering high-coverage Hi-C contact maps, enabling a more precise and comprehensive analysis of TAD structures.

\begin{figure*}[t]
\begin{center}
\includegraphics[width=1.0\linewidth]{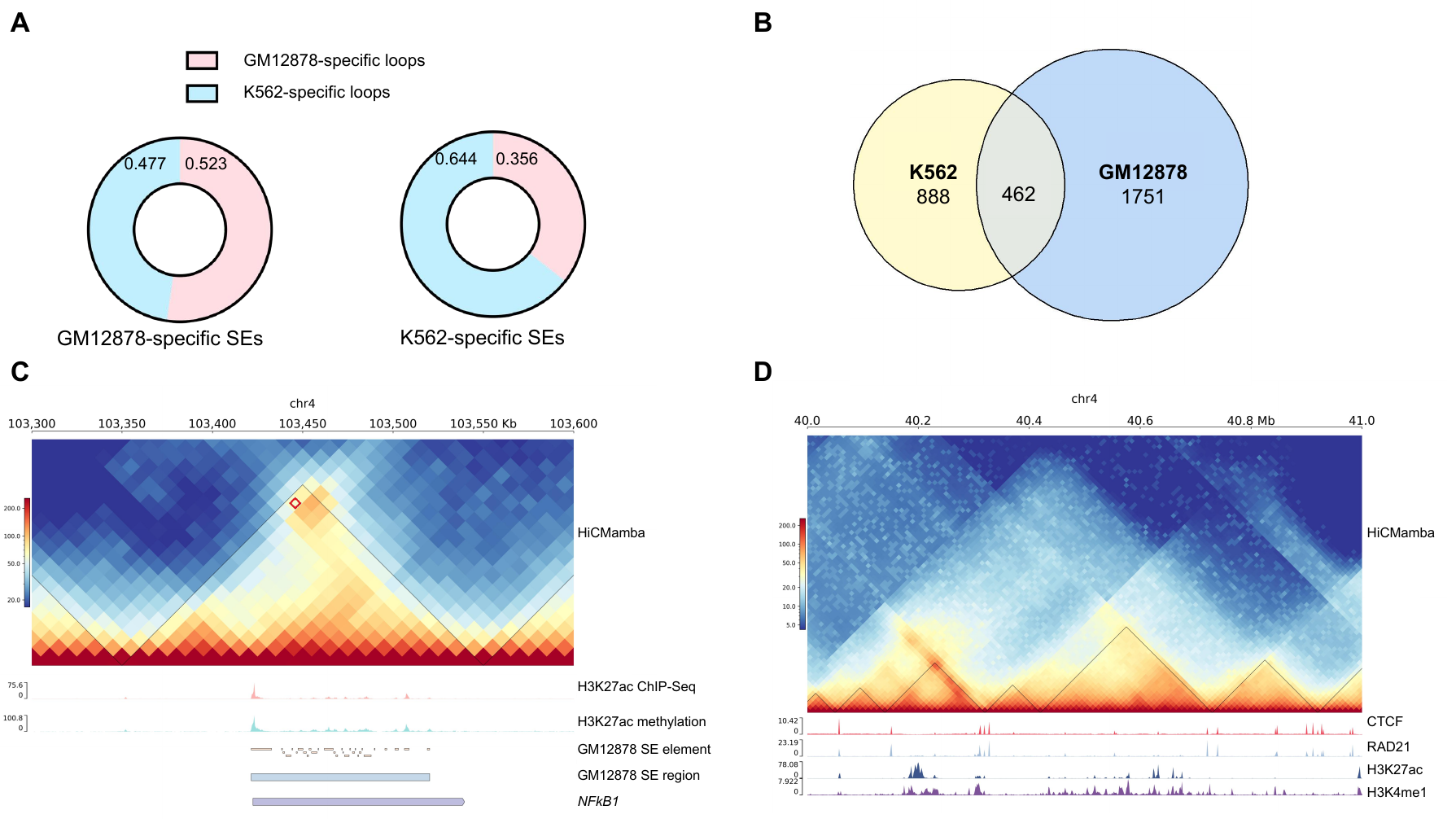}
\end{center}
\caption{
\textbf{Epigenomic features and 3D genome structures are essential for gene expression.} (\textbf{A}) Loop weighted scores of cell type-specific loops versus cell type-specific super-enhancers (SEs). (\textbf{B}) Volcano plot of differential gene expression. (\textbf{C}) Venn Diagram of Differentially Expressed Genes in GM12878 and K562 Cell Lines. (\textbf{D}) 3D genome structure and epigenomic features around the \textit{NFKB1} gene locus. (\textbf{E}) A snapshot of TAD and ChIP-Seq signals of transcription factors on chromosome 4: 40Mb-41Mb.
}
\end{figure*}

\begin{table*}
\caption{The number of cell line-specific loops associated with cell-line-specific SEs $A_\cdot^s$, the proportion of SE-related loops $P_\cdot^s$, and the loop weighted score $W_\cdot^s$. The total number of loops identified in GM12878 and K562 are $N_{\rm GM12878} = 708$ and $N_{\rm K562} = 344$, respectively.}
\setlength{\tabcolsep}{16.5mm}{
\begin{tabular}{@{}lcc@{}}
\toprule
                    & \begin{tabular}[c]{@{}c@{}}GM12878 \\ ($A_{\rm GM12878}^s$ / $P_{\rm GM12878}^s$ / $W_{\rm GM12878}^s)$ \end{tabular} & \begin{tabular}[c]{@{}c@{}}K562\\ ($A_{\rm K562}^s$ / $P_{\rm K562}^s$ / $W_{\rm K562}^s)$\end{tabular} \\ \midrule
GM12878-specific SE & 151 / 0.213/ 0.523                                                                                                                                                                 & 67 / 0.195 / 0.477                                                                                                                                                 \\
K562-specific SE    & 50 / 0.071 / 0.356                                                                                                                                                                 & 44 / 0.127 / 0.644                                                                                                                                                 \\ \bottomrule
\end{tabular}}
\end{table*}

\subsection*{Correlation between the 3D genome and epigenomics}
The organization of higher-order chromatin is essential for gene regulation and cellular homeostasis \cite{cuartero2023three, hu2023lineage}. We investigate the relationship between 3D genome structure and epigenomic features (i.e., SEs, SE elements, ChIP-Seq signals, and methylation signals), which are integral to understanding gene expression and chromatin organization. 

First, we first identify cell-type-specific SEs and loops, defining them as those not overlapping between GM12878 and K562 cell lines.
As shown in Figure~5A and Table~5, we observe a strong association between GM12878-specific SEs and loops, and similarly for K562, using a loop-weighted scoring method (detailed in the "Loop Weighted Score" section).
This findings emphasize the critical role of cell-type-specific loops in their respective functions in concordance with previous research \cite{kai2018predicting}. 

Further analysis of differentially expressed genes (DEGs) between GM12878- and K562-specific loops (Figure~5B and Figure~5C) reveals functional links to 3D genome structure.
For example, the acetylation cycle of cohesin, which modulates chromatin loop length through a \textit{PDS5A}-mediated brake mechanism \cite{van2022cohesin}, is one such process involving overlapping DEGs. Similarly, the gene \textit{DNAH3}, located adjacent to a specific deletion, exhibits chromatin interactions with enhancer elements within the deleted region \cite{jensen2021combinatorial}. These examples highlight the connection between DEGs and 3D genome organization. Moreover, cell-type-specific DEGs tend to be involved in cell differentiation. For instance, \textit{NFKB1} shows significantly higher expression in GM12878 compared to K562, with associated loops being exclusively found in GM12878. 
This observation aligns with the constitutive activation of \textit{NFKB1} pathways in the GM12878 lymphoblastoid B-cell line \cite{ang2024aberrant}.
Figure~5D depicts the loop architecture associated with \textit{NFKB1} expression, revealing concomitant increases in H3K27ac ChIP-seq and methylation peaks within the loop region.
Additionally, GM12878-specific SE and SE elements are observed around these regions, suggesting a potential regulatory role in this context. Notably, the distribution of SE elements aligns with the ChIP-Seq and methylation peaks of H3k27ac. 
These findings are corroborated by Zhao~et~al.~\cite{zhao2014nf}, who demonstrated the essential role of \textit{NFKB1} subunits, enriched at active enhancers marked by H3K27ac signals, in B cell development and function.

Finally, Figure~5E illustrates the distinct enrichment patterns of various epigenomic features within and around TADs. CTCF and RAD21 ChIP-Seq signals are enriched at the TAD boundaries, whereas the H3K27ac and H3K4me1 signals are enriched within the TADs, consistent with previous studies \cite{saha2020interplay, liu2022ctcf, sun2023rad21}. 
Altogether, HiCMamba can effectively recover chromatin interaction patterns such as loops and TADs. These 3D genome structures are intricately associated with various epigenomic features, collectively contributing to the transcriptional regulation of cell-type-specific genes.

\section*{Discussion}
Three-dimensional chromatin structures, e.g., topologically associating domains (TADs) and loops, derived from Hi-C data are essential for deciphering the intricate relationship between chromatin organization and transcription regulation. 
Obtaining high-resolution Hi-C data poses significant technical and financial challenges, leading to the prevalence of low-resolution contact maps that hinder accurate interaction frequency estimations.
In this work, we presented a UNet-based auto-encoder framework HiCMamba, empowered by the Mamba block \cite{gu2023mamba}, for the efficient and accurate in-silico enhancement of Hi-C contact maps. To the best of our knowledge, HiCMamba is the first model to harness a state space model for enhancing Hi-C resolution.

Specifically, HiCMamba combined the strengths of UNet \cite{ronneberger2015u} architecture and a novel holistic scan block to enable effective multi-scale contact map processing. The holistic scan block comprised an SS2D module, which leverages Mamba's long-range modeling capabilities for comprehensive feature extraction, and an LEFN module, which optimizes information flow for enhanced accuracy and efficiency. 
Evaluations on GM12878 and K562 Hi-C datasets demonstrated the superior performance of HiCMamba compared to the state-of-the-art deep learning methods. It achieved high-quality recovery results at a remarkably low computational cost, requiring only 25\% of the resources compared to the second-best method.
Cross-cell line experiments further validated its robust generalization capabilities.
Moreover, unlike other methods confined to local receptive fields, HiCMamba features global receptive fields, enabling the efficient representation of distant genomic loci.
Importantly, HiCMamba-enhanced contact maps yielded 3D genome structures, TADs and loops, with significantly fewer false positives compared to other methods.
Our analysis also revealed a strong association between cell-type-specific SEs and loops, highlighting the importance of these structures in cell-type-specific functions. Furthermore, we observed an intricate interplay between 3D genome organization and various epigenomic features, suggesting a combined role in regulating cell-type-specific gene expression.

Although HiCMamba demonstrated its effectiveness in Hi-C contact maps enhancement, several avenues for future improvement exist.
First, HiCMamba processes the contact maps as images, akin to an image restoration task. 
However, incorporating DNA sequence data, which has shown promise in predicting 3D genome structures \cite{fudenberg2020predicting, zhou2022sequence}, could provide a more comprehensive understanding of locus-specific contact patterns.
Additionally, while extending HiCMamba's capacity (now at 10kb resolution) to analyze higher resolutions is feasible, it would demand more fine-grained Hi-C data and increased computational resources.
This work makes a significant contribution by demonstrating the potential of state space models for resolution enhancement of Hi-C contact maps, paving the way for future advancements in the field.









\bibliography{main}

\end{document}